%% file: main.tex
\documentclass{article}

    \usepackage[final]{neurips_2024}

\usepackage[utf8]{inputenc} 
\usepackage[T1]{fontenc}    
\usepackage{hyperref}       
\usepackage{url}            
\usepackage{booktabs}       
\usepackage{amsfonts}       
\usepackage{nicefrac}       
\usepackage{microtype}      
\usepackage{xcolor}         
\usepackage{amsmath}
\usepackage{tikz}
\usepackage{listings}
\usepackage{xcolor}

\definecolor{eclipseStrings}{RGB}{34,139,34}
\definecolor{eclipseKeywords}{RGB}{220,20,60}
\colorlet{numb}{magenta!60!black}
\definecolor{lightgray}{rgb}{0.95,0.95,0.95}

\lstdefinelanguage{json}{
    basicstyle=\tiny\ttfamily,
    commentstyle=\color{eclipseStrings}, 
    stringstyle=\color{eclipseKeywords}, 
    numbers=none,
    showstringspaces=false,
    breaklines=true,
    frame=lines,
    backgroundcolor=\color{lightgray}, 
    string=[s]{"}{"},
    comment=[l]{:\ "},
    morecomment=[l]{:"},
    literate=
        *{0}{{{\color{numb}0}}}{1}
         {1}{{{\color{numb}1}}}{1}
         {2}{{{\color{numb}2}}}{1}
         {3}{{{\color{numb}3}}}{1}
         {4}{{{\color{numb}4}}}{1}
         {5}{{{\color{numb}5}}}{1}
         {6}{{{\color{numb}6}}}{1}
         {7}{{{\color{numb}7}}}{1}
         {8}{{{\color{numb}8}}}{1}
         {9}{{{\color{numb}9}}}{1}
}

\title{\textsc{Proof of Thought} : Neurosymbolic Program Synthesis allows Robust and Interpretable Reasoning}

\author{%
  Debargha Ganguly\thanks{Most work conducted  at Microsoft Research, with partial support from NSF Awards 2117439 and 2112606.} \\
  Case Western Reserve University \\
  \texttt{debargha@case.edu} \\
  \And
  Srinivasan Iyengar \\
  Microsoft Corporation \\
  \texttt{sriyengar@microsoft.com} \\
  \AND
  Vipin Chaudhary \\
  Case Western Reserve University \\
  \texttt{vipin@case.edu} \\
  \And
  Shivkumar Kalyanaraman \\
  Microsoft Corporation \\
  \texttt{shkalya@microsoft.com} \\
}

\begin{document}

\maketitle

\input{abstract}

\input{introduction}

\input{relatedworks}

\input{methodology}

\input{results}

\input{discussion}
\input{conclusion}

\bibliographystyle{plainnat} 
\bibliography{ref.bib} 

\input{appendix}

\end{document}

%% file: abstract.tex
\begin{abstract}
Large Language Models (LLMs) have revolutionized natural language processing, yet they struggle with inconsistent reasoning, particularly in novel domains and complex logical sequences. This research introduces \textsc{Proof of Thought}, a framework that enhances the reliability and transparency of LLM outputs. Our approach bridges LLM-generated ideas with formal logic verification, employing a custom interpreter to convert LLM outputs into First Order Logic constructs for theorem prover scrutiny. Central to our method is an intermediary JSON-based Domain-Specific Language, which by design balances precise logical structures with intuitive human concepts. This hybrid representation enables both rigorous validation and accessible human comprehension of LLM reasoning processes. Key contributions include a robust type system with sort management for enhanced logical integrity, explicit representation of rules for clear distinction between factual and inferential knowledge, and a flexible architecture that allows for easy extension to various domain-specific applications. We demonstrate \textsc{Proof of Thought}'s effectiveness through benchmarking on StrategyQA and a novel multimodal reasoning task, showing improved performance in open-ended scenarios. By providing verifiable and interpretable results, our technique addresses critical needs for AI system accountability and sets a foundation for human-in-the-loop oversight in high-stakes domains.
\end{abstract}

%% file: introduction.tex
\section{Introduction}
Large language models (LLMs) have revolutionized the field of AI and enabled a wide range of applications. However, as these models are increasingly deployed to process unstructured data and perform complex tasks autonomously, their inconsistent reasoning capabilities remain a critical limitation \citep{marcus2020next}. This inconsistency manifests in variable performance across out-of-domain reasoning, negation understanding, and extended logical chains, suggesting a reliance on superficial heuristics \citep{bender2021dangers}. The implications are far-reaching, particularly in high-stakes domains where reliable and transparent decision-making is crucial \citep{rudin2019stop}. Errors or biases in these contexts could have severe consequences, underscoring the urgent need for more dependable and interpretable AI systems.

Recent advances in prompt engineering have shown promise in addressing these challenges. Techniques such as Chain-of-Thought (CoT) \citep{wei2022chain}, Self-Consistency with CoT (CoT-SC) \citep{wang2022self}, Tree of Thoughts (ToT) \citep{yao2024tree}, and Graph of Thoughts (GoT) \citep{besta2024graph} have improved problem-solving capabilities. In the multimodal domain, techniques like Set-of-Marks (SoM) prompting have emerged \citep{yang2023set}. Despite advancements in performance figures, the mechanisms behind these improvements remain opaque, causing blind spots in real-world usage as failure modes are not well understood. The fundamental issue lies in the lack of interpretability and guaranteed verifiability in LLM reasoning processes \citep{danilevsky2020survey}. This opacity hinders our ability to trust and validate LLM outputs, a critical concern in scenarios requiring explainable AI or human-in-the-loop oversight.

Real-world applications, especially in health and safety domains, face additional challenges in training and deploying models due to the scarcity of high-quality annotated data. This is particularly evident in sectors like energy, healthcare, and manufacturing, where ML has significant potential for operational efficiency \citep{jordan2015machine}. Complex tasks, such as identifying OSHA violations in cluttered visual data, illustrate the difficulties in long-tail, low-data scenarios. These scenarios are characterized by diverse and unpredictable phenomena, where specialized model training is impractical \citep{zhang2021understanding}. With robust reasoning, LLMs' wide knowledge and commonsense abilities can potentially operate in these low-data paradigms, opening up more applications.

\textbf{Contributions:} To address these challenges, we :
\begin{enumerate}
    \item Propose \textsc{Proof of Thought} (PoT), a novel approach that leverages the in-context learning and code generation capabilities of LLMs while incorporating their inherent knowledge and spatial understanding. Our system employs a custom interpreter that parses "LLM-Thoughts" (represented as DSL code snippets) to generate First Order Logic programs, which are then verified by a Z3 theorem prover.
    \item Introduce an intermediate JSON-based DSL (neurosymbolic representation) and the associated interpreter that operates on human-understandable abstract concepts using intuitive, near-English language constructs (see Fig \ref{fig:architecture}). This strikes a balance between the precision required for logical definitions with formal first-order logic proofs and accessibility for non-expert users.
    \item Benchmark performance over StrategyQA (a boolean multi-hop implicit NLP reasoning benchmark) and a novel multimodal real-world long-tail reasoning problem (Reddit-OSHA Benchmark). This shows that PoT works on complex, and a wide variety of tasks.
\end{enumerate}

\textsc{Proof of Thought} enhances the capabilities of LLMs in complex, open-ended scenarios by providing reasoning guarantees, conditioned on correctness of knowledge base and rule specifications, therefore furnishing a framework for human-in-the-loop oversight and verification.

%% file: relatedworks.tex
\section{Related Work}
\textbf{Early Integration Attempts} laid the groundwork for neuro-symbolic AI. Works like EBL-ANN \citep{towell1994knowledge}, KBANN \citep{towell1990refinement}, and C-ILP \citep{d2009neural} incorporated propositional formulae into neural networks. While pioneering, these approaches struggled with scalability and expressiveness. Knowledge Graph integration advanced the field further. Methods proposed by \citet{chen2020review} and \citet{kampffmeyer2019rethinking} showed promise in leveraging structured knowledge for improved reasoning. However, maintaining interpretability and explicit rule incorporation remained challenging. Differentiable Logic Programming frameworks like DeepProbLog \citep{manhaeve2018deepproblog} and Scallop \citep{li2023scallop} demonstrated the potential of integrating probabilistic logic programming with neural networks. These approaches enable end-to-end training of neuro-symbolic systems but face limitations in handling complex reasoning tasks and diverse logical formalisms. \citet{gupta2023visual} showed how compositional visual reasoning can be done by program generation without training. 

\textbf{Large Language Models} have opened new avenues for neuro-symbolic reasoning. Techniques such as Chain-of-Thought \citep{wei2022chain}, Tree-of-Thoughts \citep{yao2024tree}, and Graph-of-Thoughts \citep{besta2024graph} have shown impressive results in complex reasoning tasks. However, these methods often produce inconsistent intermediate steps and struggle with out-of-domain reasoning. Interpretable Concept-based Models, as explored by \citet{kim2018interpretability} and \citet{chen2020concept}, aim to increase trust in deep learning models. However, state-of-the-art approaches often rely on high-dimensional concept embeddings that lack clear semantic meaning, limiting their interpretability.

\textbf{Advanced Neuro-symbolic Frameworks} like the Deep Concept Reasoner (DCR) \citep{barbiero2023interpretable} and GENOME \citep{chen2023genome} have made progress in combining neural and symbolic components. DCR constructs syntactic rule structures from concept embeddings, while GENOME introduces a modular approach for visual reasoning tasks. Despite these advancements, challenges in scalability and generalization to diverse task domains persist. A critical issue is reasoning shortcuts, where models use unintended concept semantics. To address this, \citet{marconato2024bears} introduced BEARS, improving model calibration and informative annotation acquisition. These developments showcase the potential for imparting robustness and reliability via neuro-symbolic AI systems.

\textbf{LLM Code Generation and Low-Resource Translation} demonstrate the remarkable adaptability of large language models. In code generation, LLMs like Codex \citep{chen2021evaluating} and AlphaCode \citep{li2022competition} show proficiency across multiple programming languages, often outperforming specialized models. These systems excel at understanding context, generating syntactically correct code, and even solving complex algorithmic problems. Paralleling this, research in low-resource language translation \citep{zoph2016transfer, tanzer2023benchmark} reveals LLMs' ability to rapidly adapt to new languages with minimal examples. Techniques like few-shot learning and cross-lingual transfer enable models to leverage knowledge from high-resource languages to improve performance on low-resource ones. Both domains highlight LLMs' capacity for quick adaptation and generalization, suggesting potential for enhanced neuro-symbolic systems that can efficiently learn and apply formal reasoning across diverse domains with limited training data.

%% file: methodology.tex
\section{\textsc{Proof of Thought}: A Neurosymbolic Reasoning Framework}

In this section, we introduce \textsc{Proof of Thought} (PoT), a novel framework that bridges NLP with formal logical reasoning to enhance the interpretability and verifiability of LLM outputs. We first outline the foundational concepts and notations that underpin our approach.

\begin{figure}
    \centering
    \includegraphics[width=1\linewidth]{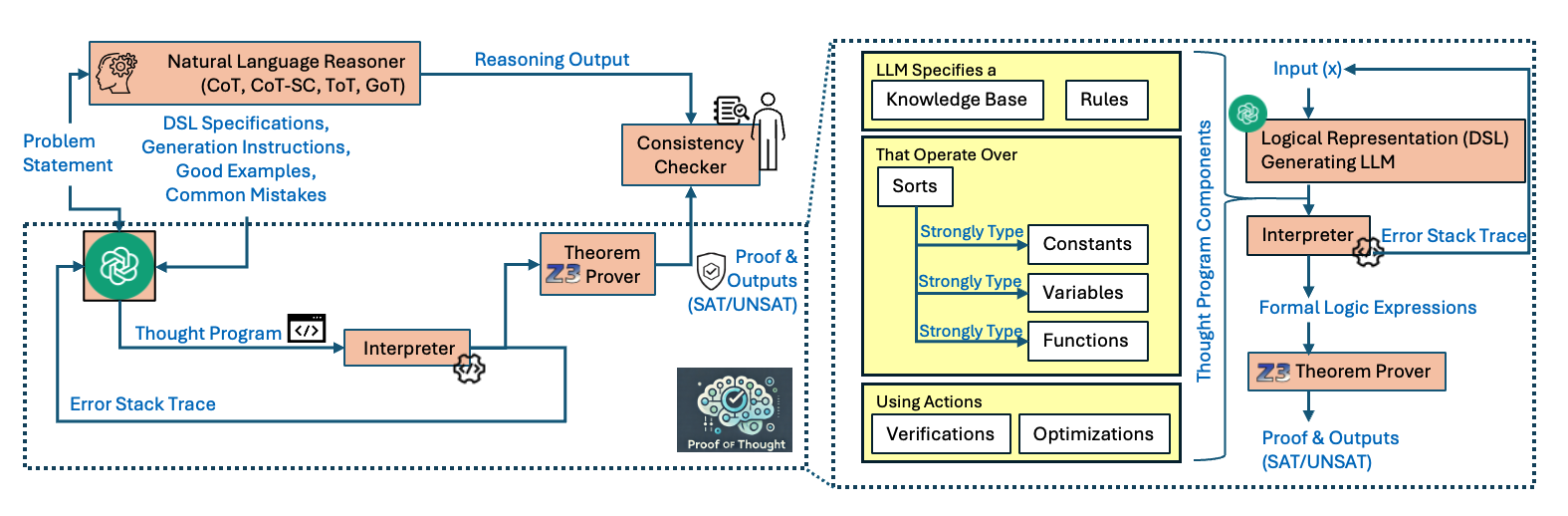}
    \caption{Architecture of the \textsc{Proof of Thought} (PoT) framework, illustrating the integration of natural language reasoning with formal logical verification.}
    \label{fig:architecture}
\end{figure}

\subsection{Background Concepts \& Notation from LLM Reasoning Literature}

Let $p_\theta$ denote a pre-trained language model (LM) with parameters $\theta$. In a conversation, user messages (prompts) and LM replies (thoughts) are exchanged. We use lowercase letters $x, y, z, ...$ to indicate LM thoughts, where the definition of a "thought" is use-case specific (e.g., a paragraph, document, or code block). The simplest Prompting Approach approach is Input-Output (IO) where an LM directly transforms an input sequence $x$ into output $y$ without intermediate steps. Chain-of-Thought (CoT) introduces intermediate thoughts $a_1, a_2, ...$  between $x$ and $y$, enhancing performance on tasks like mathematical reasoning. 

Multiple CoTs generalizes CoT by generating $k$ independent chains and selecting the best output based on a prescribed scoring metric. This approach, introduced as Self-Consistency with CoT (CoT-SC), allows exploration of different reasoning paths. Tree of Thoughts (ToT) further enhances CoT-SC by modeling reasoning as a tree of thoughts. Each node represents a partial solution. For a given node, $k$ new nodes are generated, then evaluated using an LM or human scores. The tree expansion is guided by search algorithms like BFS or DFS. Finally, Graph of Thoughts (GoT) extends ToT by allowing more complex connections between thoughts, forming a directed graph structure. In GoT, thoughts can have multiple predecessors and successors, enabling more flexible reasoning paths and the combination of ideas from different branches. This approach allows for cyclic reasoning patterns and can capture more intricate problem-solving strategies.

\subsection{Framework Overview}
\begin{quote}
  "All our knowledge begins with the senses, proceeds then to the understanding, and ends with reason. There is nothing higher than reason." 
  
  - Immanuel Kant, in \textit{Critique of Pure Reason}  
\end{quote}

\textsc{Proof of Thought} models the LLM's reasoning process as a structured transformation from natural language input to formal logical expressions that can be verified using theorem proving techniques. The framework consists of three primary components :
\begin{itemize}
\item \textbf{Logical Representation Generator} $\mathcal{G}$: Maps input $x$ to a logical representation $\mathcal{L}$ using $p_\theta$.
\item \textbf{Interpreter $\mathcal{I}$}: Parses $\mathcal{L}$ and constructs formal logical expressions $\phi$ in first-order logic (FOL).
\item \textbf{Theorem Prover $\mathcal{T}$}: Verifies the validity of $\phi$ and provides proofs or counterexamples.
\end{itemize}
The PoT reasoning process can thus be formalized as:
\[
\mathcal{L} = \mathcal{G}(x; p_\theta); \phi = \mathcal{I}(\mathcal{L}); \text{Verification Result} = \mathcal{T}(\phi)
\]

\textbf{Guarantees:} By using theorem proving, we rely on the principle of logical consequence, where conclusions are guaranteed to be true if the premises and inference rules, depicted in the logical representations (both of which are human readable, allowing interpretability and verifiability). These logical representations are the ultimate arbiter of truth and validity. The guarantees are what distinguishes PoT from other forms of reasoning. Inductive reasoning (drawing general conclusions from specific observations) can be useful, but doesn't offer the same level of certainty. In contrast, guarantees with theorem proving allow us to establish precise arguments with mathematical truths with absolute confidence, elevating the reasoning process beyond mere conjecture or intuition.

Our \textsc{Proof of Thought} framework architecture (shown in Fig \ref{fig:architecture}) introduces a JSON-based Domain Specific Language (DSL) along with the associated interpreter. Next, we discuss the design choices for these in detail.   

\subsection{Design of the JSON-Based Domain-Specific Language (DSL)}

The core of our logical reasoning system is built upon a carefully designed JSON-based Domain-Specific Language (DSL) that serves as an intermediate representation for translating reasoning tasks into formal logic. This DSL was created with the primary challenge of being general-purpose enough to accommodate a wide range of reasoning problems. The choice of JSON as the underlying format was deliberate, leveraging its widespread use, human readability, and ease of parsing. This design decision ensures that the logical representations are both machine-parseable and accessible to users who may not have expertise in formal logic or programming. Moreover, JSON's compatibility with structured outputs from AI service providers like OpenAI and Google, which offer guaranteed outputs matching particular schemas, makes it an ideal choice for our system that doesn't rely on retraining models. The key components of the DSL are:

\begin{figure}
    \centering
    \includegraphics[width=1\linewidth]{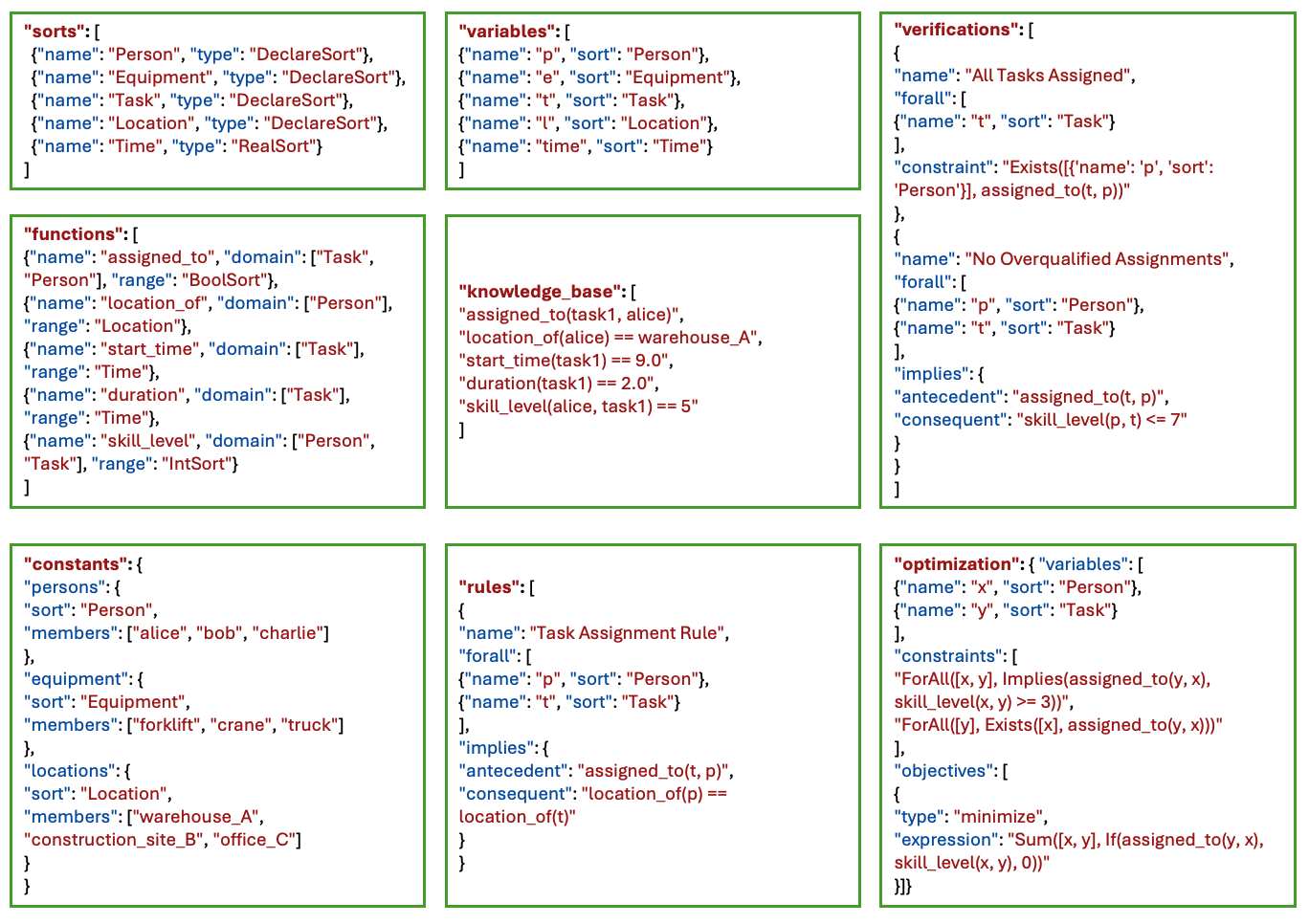}
    \caption{Example DSL program components of the \textsc{Proof of Thought} (PoT) framework for a dummy task assignment verification and optimization problem. The figure displays the JSON-based Domain-Specific Language (DSL) structure, including sort definitions, variables, functions, constants, knowledge base, rules, verifications, and optimization constraints for a workforce management scenario.}
    \label{fig:enter-label}
\end{figure}

\begin{enumerate}
    \item \textbf{Sorts ($\mathcal{S}$)} define the domains or types used in the logic. Let $\mathcal{S} = \{S_1, S_2, \ldots, S_n\}$ be the set of sorts, where each $S_i$ represents a specific domain. The inclusion of sorts in our system is a key differentiator. It allows for reasoning over high-level, human-understandable concepts, which is crucial for bridging the gap between natural language problem descriptions and formal logical representations. The sort definition in Fig \ref{fig:enter-label} allows for more intuitive problem representation. For instance, instead of reasoning about abstract entities, we can now reason about Persons, Equipment, Tasks, Locations, and Time. This makes it easier to translate natural language problems into formal logic. The type-safe reasoning catches semantic errors early. For example, if we tried to apply a function meant for Equipment to a Person, the system would catch this error before any reasoning takes place. Moreover, this structure allows for domain-specific reasoning strategies. For instance, we could implement specialized reasoning algorithms for time-based logic using the Time sort.

    \item \textbf{Functions ($\mathcal{F}$)} definitions in our system go beyond simple predicates, allowing for rich, typed relationships between sorts. Let $f : S_1 \times S_2 \times \ldots \times S_k \rightarrow S_r$ represent a function, where $S_1, S_2, \ldots, S_k, S_r \in \mathcal{S}$. For predicates, $S_r = \text{Bool}$. 
    
    For example, in Fig \ref{fig:enter-label}, the function definition allows for complex domain modeling. For instance, `assigned\_to` represents a relationship between Tasks and Persons, `location\_of` transforms a Person into a Location, and `skill\_level` represents a property of a Person in relation to a specific Task. The type-checking ensures that functions are only applied to arguments of the correct sort, reducing logical errors. For example, trying to find the `skill\_level` of an Equipment item for a Task would be caught as a type error. In future work, we hope that these function definitions also open up the possibility of incorporating external algorithms. For instance, the `duration' function could be linked to an external scheduling algorithm that calculates task durations based on various factors.

    \item \textbf{Constants ($\mathcal{C}$)} with associated sorts provide grounding for abstract reasoning in concrete entities. Let $c_i : S_j$ denote that constant $c_i$ is of sort $S_j$. In Fig \ref{fig:enter-label} constant declaration grounds the abstract sorts in concrete entities. It allows for easy integration of domain-specific knowledge - for instance, we know that "alice", "bob", and "charlie" are Persons in our system. In future work, we believe this structure has the potential for linking with external databases or knowledge graphs. For example, the "equipment" constants could be linked to an external database containing detailed specifications for each piece of equipment.    

    \item \textbf{Variables ($\mathcal{V}$)} enable clear scoping rules for quantifiers and type-safe substitutions in logical formulas. Let $x_i : S_j$ denote that variable $x_i$ ranges over sort $S_j$ (see Fig \ref{fig:enter-label}).

    \item \textbf{Knowledge Base ($\mathcal{KB}$)} contains axioms or facts assumed to be true within the logical system. Let $\mathcal{KB} = \{\varphi_1, \varphi_2, \ldots, \varphi_m\}$ where each $\varphi_i$ is a well-formed formula in first-order logic. The structured knowledge base allows for separation of axioms from rules and queries, and supports incremental knowledge addition. In Fig \ref{fig:enter-label}, the knowledge base contains factual information about the problem domain. It's separate from the rules and verifications, allowing for easy updates and additions without changing the core reasoning system. This structure also allows for potential consistency checking. For instance, we could verify that no person is assigned to two tasks at the same time, or that no task has a negative duration.

    \item \textbf{Rules ($\mathcal{R}$)} specify logical constructs, often involving quantifiers and implications. Let $\mathcal{R} = \{r_1, r_2, \ldots, r_l\}$ where each $r_i$ is a well-formed formula representing a rule. The explicit representation of rules enables a clear distinction between factual knowledge and inferential knowledge. These rules represent inferential knowledge - knowledge that can be derived from the facts in the knowledge base. They allow for easy addition of domain-specific reasoning patterns. This explicit representation of rules also supports explainable AI. When the system makes an inference, it can point to the specific rule(s) used, making its reasoning process transparent and understandable to users. For example, some rules are depicted in Fig \ref{fig:enter-label} and here are a few more:
    \begin{itemize}
        \item $\forall x : \text{Person}, \text{Worker}(x) \rightarrow \exists y : \text{Equipment}, \text{Wearing}(x, y)$
        \item $\forall x : \text{Person}, \forall y : \text{Equipment}, \text{Wearing}(x, y) \land \text{SafetyGear}(y) \rightarrow \text{Safe}(x)$
    \end{itemize}

    \item \textbf{Verifications ($\mathcal{V}$)} state properties or conditions to be verified by the theorem prover. Let $\mathcal{V} = \{v_1, v_2, \ldots, v_p\}$ where each $v_i$ is a well-formed formula to be verified. Separating verifications from the knowledge base and rules allows for clear goal-directed reasoning. These verifications (Fig \ref{fig:enter-label}) represent specific properties we want to check in our system. They allow for easy testing and validation of the knowledge base and rules. When a verification fails, the system can potentially generate counter-examples, providing valuable insights into why the desired property doesn't hold.  For example:
    \begin{itemize}
        \item $\forall x : \text{Person}, \text{Worker}(x) \rightarrow \text{Safe}(x)$
        \item $\exists x : \text{Person}, \text{Worker}(x) \land \lnot \text{Safe}(x)$
    \end{itemize}

    \item \textbf{Optimization (optional) ($\mathcal{O}$)} sections define  problems with objectives and constraints. Let $\mathcal{O} = (f_\text{obj}, \mathcal{C})$ where $f_\text{obj}$ is the objective function and $\mathcal{C}$ is a set of constraints. The optimization problem in Fig \ref{fig:enter-label} seeks to minimize the total skill level of assigned persons while ensuring that all tasks are assigned and that each assigned person has at least the minimum required skill level. This ability to combine logical constraints with numerical optimization allows for the representation of complex real-world problems that go beyond pure logical satisfiability. We intend to benchmark this in future work. Here is an example:
    \begin{itemize}
        \item $\text{minimize } f_\text{obj}(x) = \sum_{i=1}^n \text{cost}(x_i)$
        \item subject to: $\forall i, 1 \leq i \leq n, \text{Safe}(x_i)$
    \end{itemize} 

    \item \textbf{Actions ($\mathcal{A}$)} list what the interpreter has to perform. The only two possible actions are `verify' and `optimize'. This simple declaration tells the system what to do with the problem representation we've built up. It provides flexibility in choosing different reasoning or optimization approaches for the same problem representation. For extensibility, this structure also opens up possibilities for meta-reasoning about which actions to take. For instance, the system could analyze the problem structure to decide whether to attempt verification first or go straight to optimization.

\end{enumerate}

\subsection{Design of the Interpreter's Facilities and Capabilities}

This section provides an in-depth description of the interpreter's facilities, detailing how it constructs and manipulates logical expressions. 

\textbf{Type System, Sort Management: }The interpreter implements a robust type system, managing sorts and ensuring type safety across all expressions. It supports a variety of Z3 compatible sorts, including primitive sorts like Bool, Int, and Real, which form the foundation of the type system. User-defined sorts, known as declared sorts, allow for the representation of specific domains such as Person or Equipment. For situations requiring a finite set of elements, enumerated sorts are available. The type system also accommodates composite sorts, constructed using type constructors, which enable the creation of function sorts or tuple sorts. Throughout its operations, the interpreter rigorously enforces type consistency, ensuring that functions and predicates are applied only to arguments of the correct sorts.

\textbf{Symbol Table, Scope Management: }Central to the interpreter's functionality is a symbol table that maintains mappings from identifiers to their definitions, including variables, constants, and functions. This table is crucial for scope management, particularly when dealing with quantified variables in logical expressions. The parsing process is another key component, where the interpreter builds abstract syntax trees (ASTs) that represent the structure of expressions. This process handles a wide range of logical constructs, from atomic formulas (basic predicates applied to terms) to complex formulas constructed using logical connectives and quantifiers. The interpreter pays special attention to quantifiers, carefully managing bound variables to ensure correct scoping. Additionally, it supports substitution of terms for variables, an essential operation in applying inference rules.

\textbf{Pre-processing: }While the bulk of reasoning is handled by the theorem prover, the interpreter applies basic inference and simplification rules to optimize expressions before passing them on. This includes simplification processes that reduce expressions using logical identities, such as eliminating double negations. Normalization is another crucial step, converting expressions into a standard form (like prenex normal form) to facilitate theorem proving. The interpreter also performs early error detection, identifying contradictory statements or type mismatches before they can cause issues in later stages of processing.

\textbf{Feedback Loop:} Adequate error handling and diagnostics are paramount in the interpreter's design. It provides detailed error messages to assist the LLM in identifying and correcting issues with its programs. These diagnostics cover a range of potential problems, including type errors that indicate inconsistencies or mismatches in the type system, alerts for undefined symbols when functions, predicates, or constants are used without proper definition, and syntax errors that highlight issues in the structure of logical expressions.

\textbf{Future Proofing: }The interpreter's architecture emphasizes extensibility and customization. Users have the flexibility to extend its capabilities in several ways. They can add new sorts to define additional domains of discourse, expanding the system's ability to represent complex scenarios. The logical language can be enhanced by defining new functions and predicates, allowing users to capture more intricate relationships within their domain of interest. Furthermore, the modular design of the interpreter facilitates integration with different theorem provers or logic systems, enhancing its versatility and applicability to various problem domains.

%% file: results.tex
\section{Results}

\subsection{StrategyQA - Complex Natural Language Reasoning}
\textbf{Task Setup:} StrategyQA presents a significant challenge in natural language processing, testing a model's ability to perform multi-hop, implicit reasoning across diverse scenarios. This boolean question answering benchmark requires models to infer unstated reasoning steps, mirroring complex human cognitive processes. For example, "Did Aristotle use a laptop?" requires the implicit chain: "When did Aristotle live? When was the laptop invented? Do these time periods overlap?" This level of abstraction surpasses simple fact retrieval or explicit reasoning tasks in other benchmarks like BoolQ or Twenty Questions (20QA). While state-of-the-art language models have shown impressive performance on StrategyQA (e.g., PaLM-2 achieving 90.20\% accuracy with few-shot Chain of Thought and Self Consistency \citep{anil2023palm}), they lack transparency and verifiability in their reasoning process. Our \textsc{Proof of Thought} (PoT) framework addresses this limitation by providing complete, explicit, and verifiable reasoning chains. PoT breaks down implicit reasoning steps into explicit logical representations, defines the knowledge base used, and ensures each inference is provable through a theorem prover.

\textbf{PoT Results:} We evaluated PoT on a sample of 1000 questions from the StrategyQA dataset, focusing on the framework's ability to generate syntactically correct programs, produce provable reasoning chains, and match outputs with correct answers. The system, with the inclusion of a 3 step feedback loops (i.e., initial prompt, +2 attempts at resolving), successfully compiled and executed 82.4\% of the 1000 processed questions, marking a significant improvement from runs with lower feedback loops. This increase in compilation success rate underscores the effectiveness of our feedback mechanism in addressing and resolving issues in generated logical representations.
The system demonstrated strong recall at 91.40\%, indicating its proficiency in identifying true positive cases. The F1 score of 71.13\% suggests a good balance between precision and recall, though precision (58.22\%) presents an area for potential enhancement. The high recall, coupled with a false positive rate of 53.98\%, indicates a tendency for the system to overpredict positive cases. This observation points to a need for future refinements in discriminating between positive and negative instances more accurately.
While the compilation success rate is encouraging, the 17.6\% of questions that failed to compile highlight an area for further improvement. Enhancing the robustness of the code generation process through improved prompting techniques, fine-tuning, and expansion of the feedback loop mechanism could potentially reduce this failure rate in future iterations.

\subsection{Multimodal Reddit-OSHA Benchmark}

\textbf{Task Setup:} We curated 103 samples from the r/OSHA subreddit, featuring individuals in extremely hazardous situations. This dataset represents long-tail, low-probability scenarios, mirroring challenging real-world deployment conditions for health and safety applications. The images encompass a wide range of problems with varied lighting, scene setups, visual clutter, and resolutions.

\textbf{Baselines:} We implemented four reasoning strategies using GPT-4 as the underlying language model: Chain of Thought (CoT), Chain of Thought with Self-Consistency (CoT-SC), Tree of Thought (ToT), and Graph of Thought (GoT). Each baseline processes base64-encoded images and uses a consistent system prompt instructing the model to act as a safety inspector. CoT encourages step-by-step reasoning, concluding with a binary hazard decision. CoT-SC extends this by generating 5 independent reasoning paths per image, with the final decision determined by majority voting. ToT explores multiple reasoning paths in a tree-like structure with a maximum depth of 3 and a breadth of 2 at each node. GoT implements an iterative reasoning process with 3 iterations per image, building upon previous analyses. Evaluation metrics include win rate (proportion of correctly identified hazards) and reasoning richness (number of sentences in model responses).

\textbf{Baseline Results:} All reasoning strategies demonstrated high performance, with CoT achieving a 99.03\% win rate and CoT-SC, ToT, and GoT all achieving perfect 100\% win rates. This suggests that advanced reasoning strategies can correct errors made by simpler approaches. Reasoning richness varied significantly, with ToT producing the most detailed responses (often over 1000 sentences per image) and CoT the most concise (25-30 sentences).

\textbf{PoT Results:} Our \textsc{Proof of Thought} framework showed remarkable improvements on this dataset with the inclusion of a 3 step feedback loop (i.e., initial prompt, +2 attempts at resolving). Notably, we reduced compilation errors from 14.6\% to 0\%, demonstrating the effectiveness of our feedback and error correction mechanisms. The win rate on compiled programs increased from 72\% to 81.55\%, indicating both more reliable code generation and more accurate logical reasoning.

%% file: conclusion.tex
\section{Discussion and Future work}

Future research directions include expanding PoT to handle more complex logical structures, that extend past boolean SAT \& UNSAT, using one versus all setups, and developing more sophisticated feedback mechanisms to further reduce compilation errors. We intend to explore JSON-like representations for non-boolean responses. Additionally, exploring ways to make the logical representations more accessible to non-expert users and investigating the scalability of PoT to larger, more diverse datasets will be important next steps. Further, integrating PoT with other techniques such as model updates using reinforcement learning, or supervised fine-tuned models on synthetically generated syntactically correct PoT programs might unlock at-scale ``System 2'' thinking.

\section{Conclusion}

\textsc{Proof of Thought} bridges the gap between language models' flexibility and formal logic's rigor, offering a promising solution for trustworthy reasoning in vision-language models. By enhancing interpretability and providing reasoning guarantees, PoT addresses critical challenges in AI system accountability and reliability. Our results demonstrate its potential in both natural language and multimodal reasoning tasks, paving the way for more transparent, verifiable AI systems capable of complex reasoning in high-stakes domains.

%% file: appendix.tex
\newpage

\section{Appendix}

\subsection{StrategyQA Results}
\begin{figure}[!htbp]
    \centering
    \includegraphics[width=1\linewidth]{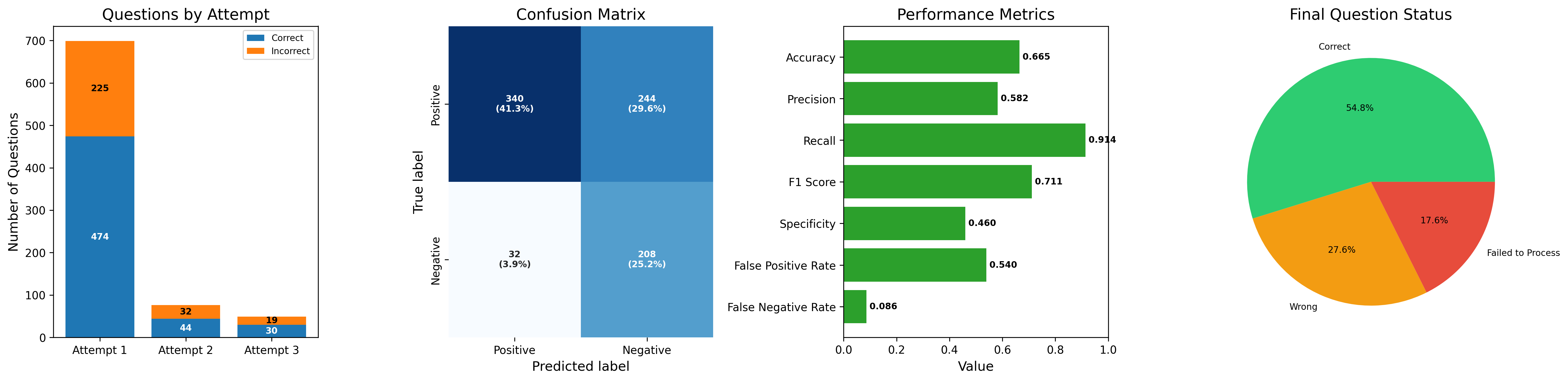}
    \caption{Performance analysis of the Proof of Thought (PoT) framework on the StrategyQA dataset. The includes four visualizations: (1) a stacked bar chart showing questions answered by attempt, (2) a pie chart displaying the final question status, (3) a confusion matrix for predicted vs. true labels, and (4) a bar chart of various performance metrics including accuracy, precision, recall, F1-score, specificity, and false positive rate.}
    \label{fig:strategyqa}
\end{figure}

\subsection{Multimodal Reddit-OSHA Benchmark}
\begin{figure}[!htbp]
    \centering
    \includegraphics[width=1\linewidth]{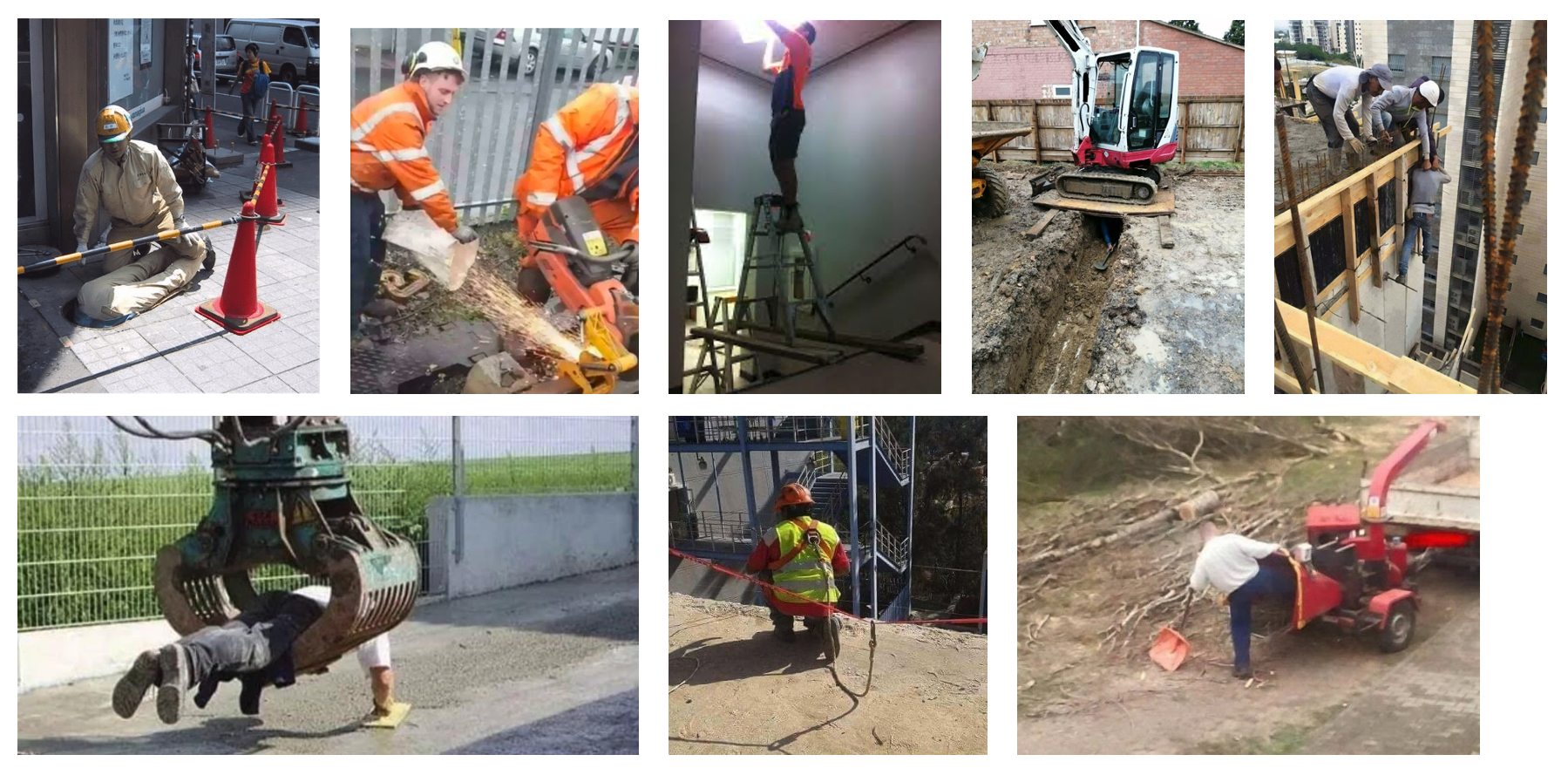}
    \caption{Sample Images from the Multimodal Reddit-OSHA Benchmark}
    \label{fig:multimodal}
\end{figure}
\begin{figure}
    \centering
    \includegraphics[width=0.6\linewidth]{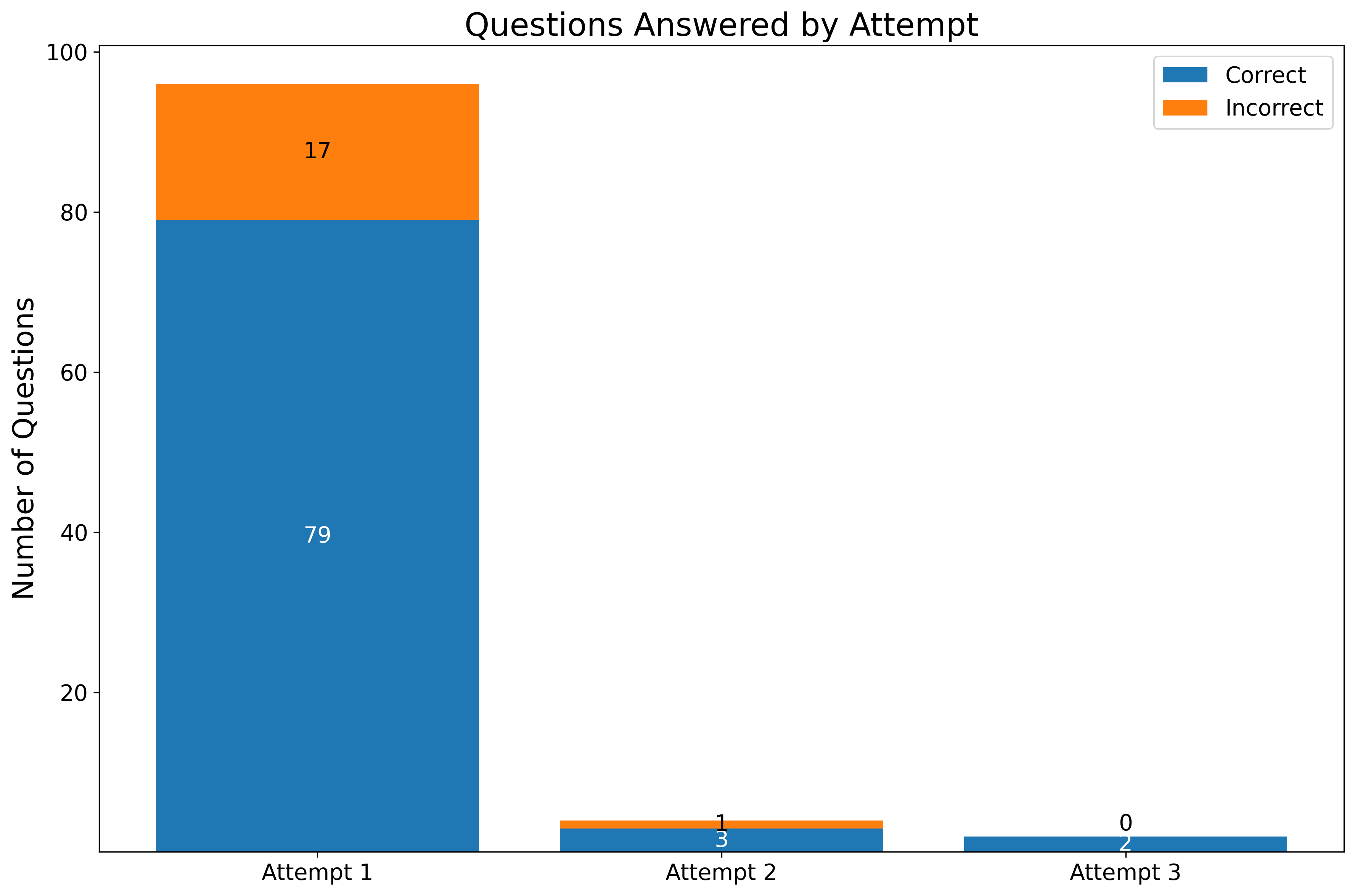}
    \caption{ Performance analysis of the Proof of Thought (PoT) framework on the Multimodal Reddit-OSHA Benchmark dataset with a stacked bar chart showing questions answered by attempt.}
    \label{fig:osha}
\end{figure}

\subsection{Real Reasoning Traces using PoT : StrategyQA Dataset}

\begin{enumerate}    

\item     \textbf{Question: } Could Javier Sotomayor jump over the head of the average giraffe? \newline \textbf{Answer: } False \newline 
\textbf{Predicted Answer: } UNSAT (False).

\begin{lstlisting}[language=json,firstnumber=1]

{
    "sorts": [
        {"name": "Person", "type": "DeclareSort"},
        {"name": "Animal","type": "DeclareSort"},
        {"name": "Real","type": "RealSort"}
    ],
    "functions": [
        {"name": "jump_height","domain": ["Person"],"range": "Real"},
        {"name": "height","domain": ["Animal"],"range": "Real"}
    ],
    "constants": {
        "persons": {"sort": "Person","members": ["javier_sotomayor"]},
        "animals": {"sort": "Animal","members": ["average_giraffe"]}
    },
    "variables": [
        {"name": "p","sort": "Person"},
        {"name": "a","sort": "Animal"}
    ],
    "knowledge_base": [
        {"assertion": "jump_height(javier_sotomayor) == 2.45"},
        {"assertion": "height(average_giraffe) == 5.5"}
    ],
    "verifications": [
        {"name": "Sotomayor Jump Over Giraffe","constraint": "jump_height(javier_sotomayor) >= height(average_giraffe)"
        }
    ],
    "actions": ["verify_conditions"]
}
\end{lstlisting}
\item     Question: Did the Cherokee people send a delegation to oppose allotment? \newline
\textbf{Answer}: True \newline
\textbf{Predicted Answer}: SAT (True)
\begin{lstlisting}[language=json,firstnumber=1]
{
    "sorts": [
        {"name": "Group","type": "DeclareSort"},
        {"name": "Action","type": "DeclareSort"},
        {"name": "Bool","type": "BoolSort"}
    ],
    "functions": [
        {"name": "send_delegation","domain": ["Group"],"range": "Bool"},
        {"name": "oppose_allotment","domain": ["Group"],"range": "Bool"}
    ],
    "constants": {
        "groups": {"sort": "Group","members": ["cherokee_people"]},
        "actions": {"sort": "Action","members": ["allotment"]}
    },
    "variables": [
        {"name": "g","sort": "Group"}
    ],
    "knowledge_base": [
        {"assertion": "send_delegation(cherokee_people)"},
        {"assertion": "ForAll([g], Implies(send_delegation(g), oppose_allotment(g)))",
            "variables": [{"name": "g","sort": "Group"}]
        }
    ],
    "verifications": [
        {"name": "Cherokee Oppose Allotment","constraint": "oppose_allotment(cherokee_people)"}
    ],
    "actions": ["verify_conditions"]
}
\end{lstlisting}

\end{enumerate}

\subsection{Real Reasoning Traces using PoT : Multimodal Reddit-OSHA Dataset}

\begin{figure}[!htbp]

\centering
\includegraphics[width=.3\textwidth]{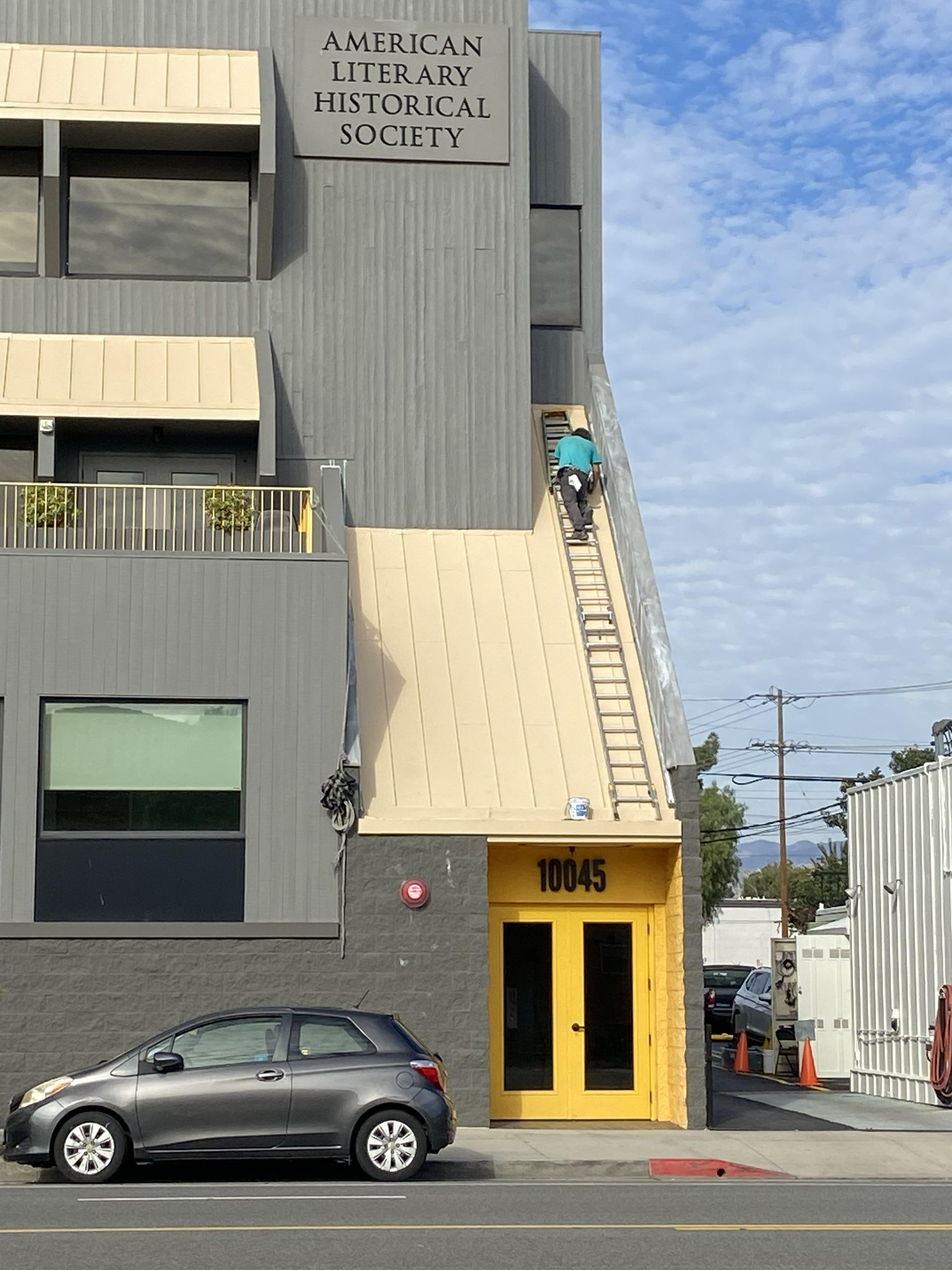}\hfill
\includegraphics[width=.23\textwidth]{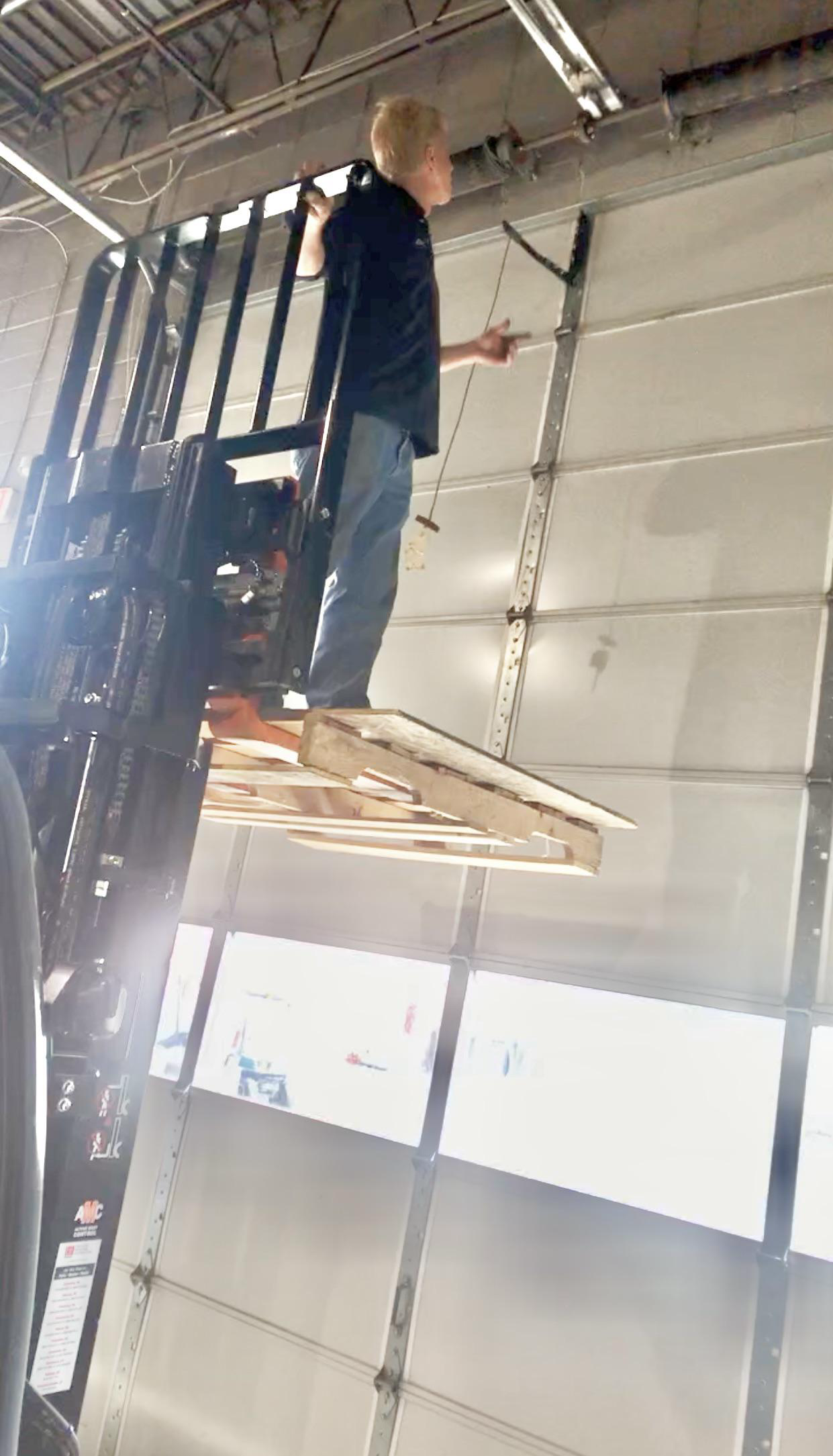}\hfill
\includegraphics[width=.3\textwidth]{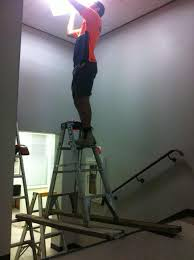}

\caption{Example 1-3 from the HSE Reddit OSHA Dataset.}
\label{fig:figure3}

\end{figure}

\begin{enumerate}
    \item HSE Example 1
    \begin{lstlisting}[basicstyle=\tiny, language=json,firstnumber=1]

{
    "sorts": [
        {"name": "Person","type": "DeclareSort"},
        {"name": "Equipment","type": "DeclareSort"},
        {"name": "SafetyGear","type": "DeclareSort"}
    ],
    "functions": [
        {"name": "Using","domain": ["Person","Equipment"],"range": "BoolSort"
        },
        {"name": "Wearing","domain": ["Person","SafetyGear"],"range": "BoolSort"}
    ],
    "constants": {
        "persons": {"sort": "Person","members": ["worker"]},
        "equipments": {"sort": "Equipment","members": ["ladder"]},
        "safetyGears": {"sort": "SafetyGear","members": ["hardHat","harness"]}
    },
    "knowledge_base": [
        {"assertion": "Using(worker, ladder)","value": true},
        {"assertion": "Wearing(worker, hardHat)","value": false},
        {"assertion": "Wearing(worker, harness)","value": false}
    ],
    "rules": [
        {"name": "Hard Hat Rule","forall": [{"name": "p","sort": "Person"},
        {"name": "e","sort": "Equipment"}],"implies": {"antecedent": "Using(p, e)", "consequent": "Wearing(p, hardHat)"}},
        {"name": "Harness Rule","forall": [{"name": "p","sort": "Person"}, {"name": "e","sort": "Equipment"}],
            "implies": {"antecedent": "Using(p, e)","consequent": "Wearing(p, harness)"}}
    ],
    "verifications": [
        {"name": "Verify Hard Hat Compliance","constraint": "Wearing(worker, hardHat)"},
        {"name": "Verify Harness Compliance","constraint": "Wearing(worker, harness)"}
    ],
    "actions": ["verify_conditions"]
}
\end{lstlisting}

\item HSE Example 2
    \begin{lstlisting}[basicstyle=\tiny, language=json,firstnumber=1]

{
    "sorts": [{"name": "Person","type": "DeclareSort"},
        {"name": "Equipment","type": "DeclareSort"},
        {"name": "SafetyEquipment","type": "DeclareSort"}
    ],
    "functions": [
        {"name": "StandingOn","domain": ["Person","Equipment"],"range": "BoolSort"},
        {"name": "UsingSafetyEquipment","domain": ["Person","SafetyEquipment"],"range": "BoolSort"},
        {"name": "IsSafe","domain": ["Person"],"range": "BoolSort"}
    ],
    "constants": {
        "persons": {"sort": "Person","members": ["worker"]},
        "equipments": {"sort": "Equipment","members": ["forklift","pallet"]},
        "safetyEquipments": {"sort": "SafetyEquipment","members": ["harness"]}
    },
    "knowledgebase": [
        {"assertion": "StandingOn(worker, pallet)","value": true},
        {"assertion": "UsingSafetyEquipment(worker, harness)","value": false}
    ],
    "rules": [
        {"name": "Safety Rule","forall": [{"name": "p","sort": "Person"}],
            "implies": {"antecedent": "And(StandingOn(p, pallet), Not(UsingSafetyEquipment(p, harness)))","consequent": "Not(IsSafe(p))"}}
    ],
    "verifications": [
        {"name": "Verify Safety","constraint": "IsSafe(worker)"}
    ],
    "actions": ["verify_conditions"]
}
\end{lstlisting}

\item HSE Example 3
    \begin{lstlisting}[basicstyle=\tiny, language=json,firstnumber=1]

{
    "sorts": [
        {"name": "Person","type": "DeclareSort"},
        {"name": "Equipment","type": "DeclareSort"},
        {"name": "Location","type": "DeclareSort"}
    ],
    "functions": [
        {"name": "Worker","domain": ["Person"],"range": "BoolSort"},
        {"name": "Using","domain": ["Person","Equipment"],"range": "BoolSort"},
        {"name": "AtHeight","domain": ["Person"],"range": "BoolSort"},
        {"name": "HasFallProtection","domain": ["Person"],"range": "BoolSort"},
        {"name": "Stable","domain": ["Equipment"],"range": "BoolSort"}
    ],
    "constants": {
        "persons": {"sort": "Person","members": ["worker"]},
        "equipments": {"sort": "Equipment","members": ["ladder","scaffold"]},
        "locations": {"sort": "Location","members": ["worksite"]}
    },
    "knowledge_base": [
        "Worker(worker)","Using(worker, ladder)","Using(worker, scaffold)", "AtHeight(worker)",
        {"assertion": "Stable(ladder)","value": false},
        {"assertion": "Stable(scaffold)","value": false},
        {"assertion": "HasFallProtection(worker)","value": false}
    ],
    "rules": [
        {
        "name": "Safety Rule","forall": [{"name": "p","sort": "Person"}],
            "implies": {"antecedent": "And(Worker(p), AtHeight(p))", "consequent": "HasFallProtection(p)"}
        },
        {
            "name": "Stability Rule","forall": [{"name": "e","sort": "Equipment"}],
            "implies": {"antecedent": "Using(worker, e)","consequent": "Stable(e)"}
        }],
    "verifications": [
        {"name": "Verify Safety", "constraint": "And(HasFallProtection(worker), Stable(ladder), Stable(scaffold))"}
    ],
    "actions": ["verify_conditions"]
}
\end{lstlisting}

\end{enumerate}

\subsection{A Qualitative Analysis of Generated DSL Programs and Reasoning Patterns}

The generated DSL programs across both the StrategyQA dataset and the OSHA dataset illustrate how formal logical representations can be used to model complex reasoning tasks. In both cases, the structured use of sorts, functions, rules, and verifications ensures that the questions posed are systematically decomposed into logical assertions that can be verified by a theorem prover such as Z3. Here, we analyze key aspects of these programs and how they contribute to effective reasoning.

\subsubsection{\textbf{Sorts and Function Definitions as the Backbone of Logical Modeling}}

The use of \verb|DeclareSort|, \verb|BoolSort|, \verb|RealSort|, and other basic sorts in the DSL programs serves as the foundation for defining the domains of discourse. For example, in the StrategyQA question involving Javier Sotomayor and a giraffe, the \verb|Person| and \verb|Animal| sorts allow the definition of relationships between humans and animals in terms of measurable attributes (e.g., jump height and height). Similarly, in the OSHA-related examples, \verb|Person|, \verb|Equipment|, and \verb|SafetyGear| sorts model the entities relevant to workplace safety.

By defining functions like \verb|jump_height|, \verb|height|, \verb|Wearing|, and \verb|Using|, we map the relationships between entities and their properties. These functions serve as predicates that are later used in verifications or rule implications. In these cases, the functions provide critical context for understanding the state of the world and the conditions under which certain outcomes (e.g., compliance with safety regulations or reaching a height) hold true.

\subsubsection{\textbf{Knowledge Base and Its Role in Establishing Ground Truth}}

The \verb|knowledge_base| section plays a vital role in grounding the reasoning process by introducing factual information, such as the jump height of Javier Sotomayor (2.45 meters) or the height of an average giraffe (5.5 meters). This knowledge is essential for theorem proving because it establishes the foundational truths that the logical system will work with. Similarly, in the OSHA examples, the knowledge base specifies whether a worker is using certain equipment, wearing protective gear, or working at height.

In both datasets, the knowledge base is used to capture the known facts that are assumed to be true at the start of reasoning. This helps set the initial conditions for the logical rules to be applied.

\subsubsection{\textbf{Rules as Key Drivers of Logical Implication}}

The \verb|rules| section formalizes the relationships between entities based on conditional logic. These rules encapsulate the domain knowledge and drive the reasoning process. For instance, the "Hard Hat Rule" in the OSHA examples states that if a person is using equipment, they should also be wearing a hard hat, while the "Harness Rule" mandates that a harness should be worn when using certain equipment. In the StrategyQA example, no explicit rules are needed beyond the basic comparison of heights.

These rules introduce a level of generalization that allows the reasoning process to handle not just specific instances but also classes of entities. For example, the rule:

\begin{lstlisting}[basicstyle=\tiny, language=json,firstnumber=1]
    {
    "name": "Hard Hat Rule",
    "forall": [{"name": "p","sort": "Person"}, {"name": "e","sort": "Equipment"}], 
    "implies": {"antecedent": "Using(p, e)", "consequent": "Wearing(p, hardHat)"}
    }
\end{lstlisting}
is applicable to any person and any equipment. This generalization is key in enabling logical inferences beyond the immediate facts provided in the knowledge base.

\subsubsection{\textbf{Verifications as the Core of Decision-Making}}

The \verb|verifications| section in these programs forms the basis of decision-making by checking whether the conditions specified in the knowledge base and rules hold. In the StrategyQA case, the verification checks if Javier Sotomayor's jump height is greater than or equal to the height of an average giraffe. The outcome of this check (UNSAT) indicates that it is false, thus the predicted answer matches the correct answer.

In the OSHA-related examples, verifications are used to ensure compliance with safety rules. For example, the program checks if the worker is wearing a hard hat or harness while using a ladder. These verifications serve as the final step in determining whether the conditions needed for safety are met or not. The results of these verifications provide a direct SAT (true) or UNSAT (false) outcome, which can then be used to assess compliance or answer the given question.

\subsubsection{\textbf{Patterns in Error Detection and Resolution}}

The DSL framework helps detect inconsistencies or non-compliance in the input data. For example, the OSHA programs can flag situations where workers are using unsafe equipment or failing to follow safety protocols. This is achieved by systematically comparing the facts provided in the knowledge base with the rules and verifications, ensuring that errors are caught before any conclusions are drawn.

Additionally, the explicit use of logical operators like \verb|And|, \verb|Not|, and \verb|Implies| makes it easy to trace the reasoning path when an outcome is SAT (true) or UNSAT (false). This traceability allows users to understand why a particular result was obtained, making the reasoning process more transparent and interpretable.

\subsubsection{Overall Analysis and Utility of Generated DSL Programs}

Across both the StrategyQA and OSHA datasets, the use of DSL programs enables structured, logical reasoning that is verifiable and interpretable. The modularity of the DSL—where different aspects of the reasoning process (entities, relationships, rules, and verifications) are clearly separated—ensures that the programs remain adaptable to a variety of problem domains.

The reasoning traces provided by these DSL programs offer significant benefits:

\begin{itemize}
    \item \textbf{Interpretability}: The modular structure makes it easy to follow the logical steps leading to a conclusion.
    \item \textbf{Error Detection}: The use of formal logic allows for early detection of contradictions or violations of safety rules.
    \item \textbf{Scalability}: The DSL framework can handle increasingly complex scenarios by adding new sorts, functions, rules, and verifications.
    \item \textbf{Generalization}: Rules written in a generalized form (e.g., using \verb|ForAll|) can be applied across different entities and scenarios, making the system more flexible.
\end{itemize}

\subsection{Exploring the possibilities: Satisfiable Neurosymbolic Programs}
Our DSL designed to be very expressive, and future proof for additional scenarios. In this subsection we present some example problems that can be expressed and solved to be found SAT.
\begin{enumerate}

\item \textbf{Simple Arithmetic Verification} : Verify that there exists an integer x such that x + 2 = 5.
    \begin{lstlisting}[basicstyle=\tiny, language=json,firstnumber=1]

{
  "sorts": [
    {"name": "Int", "type": "IntSort"}
  ],
  "functions": [],
  "constants": {},
  "knowledge_base": [],
  "rules": [],
  "verifications": [
    {
      "name": "verify_addition",
      "exists": [
        {"name": "x", "sort": "Int"}
      ],
      "constraint": "x + 2 == 5"
    }
  ],
  "actions": ["verify_conditions"]
}

\end{lstlisting}

\item \textbf{Basic Safety Equipment Rule} :  Ensure all workers are wearing hard hats.
    \begin{lstlisting}[basicstyle=\tiny, language=json,firstnumber=1]

{
    "sorts": [
      {"name": "Person", "type": "DeclareSort"},
      {"name": "Equipment", "type": "DeclareSort"}
    ],
    "functions": [
      {"name": "Worker", "domain": ["Person"], "range": "BoolSort"},
      {"name": "Wearing", "domain": ["Person", "Equipment"], "range": "BoolSort"}
    ],
    "constants": {
      "persons": {
        "sort": "Person",
        "members": ["alice", "bob"]
      },
      "equipments": {
        "sort": "Equipment",
        "members": ["hardHat"]
      }
    },
    "knowledge_base": [
      "Worker(alice)",
      "Worker(bob)",
      "Wearing(alice, hardHat)"
    ],
    "rules": [
      {
        "name": "Hard Hat Rule",
        "forall": [
          {"name": "p", "sort": "Person"}
        ],
        "implies": {
          "antecedent": "Worker(p)",
          "consequent": "Wearing(p, hardHat)"
        }
      }
    ],
    "verifications": [
      {
        "name": "Check Hard Hat Compliance",
        "forall": [
          {"name": "p", "sort": "Person"}
        ],
        "implies": {
          "antecedent": "Worker(p)",
          "consequent": "Wearing(p, hardHat)"
        }
      }
    ],
    "actions": ["verify_conditions"]
  }
\end{lstlisting}

  \item \textbf{Parent-Child Relationship} : Define a family tree and verify that a grandparent relationship holds.
     \begin{lstlisting}[basicstyle=\tiny, language=json,firstnumber=1]
 
  {
    "sorts": [
      {"name": "Person", "type": "DeclareSort"}
    ],
    "functions": [
      {"name": "parent_of", "domain": ["Person"], "range": "Person"}
    ],
    "constants": {
      "persons": {
        "sort": "Person",
        "members": ["alice", "bob", "charlie"]
      }
    },
    "knowledge_base": [
      "parent_of(bob) == alice",
      "parent_of(charlie) == bob"
    ],
    "rules": [],
    "verifications": [
      {
        "name": "Verify Grandparent",
        "constraint": "parent_of(parent_of(charlie)) == alice"
      }
    ],
    "actions": ["verify_conditions"]
  }

\end{lstlisting}

  \item \textbf{Transitive Relation Verification} : Verify that a transitive property holds in a relation.
    \begin{lstlisting}[basicstyle=\tiny, language=json,firstnumber=1]

{
    "sorts": [
      {"name": "Element", "type": "DeclareSort"}
    ],
    "functions": [
      {"name": "related", "domain": ["Element", "Element"], "range": "BoolSort"}
    ],
    "constants": {
      "elements": {
        "sort": "Element",
        "members": ["x", "y", "z"]
      }
    },
    "knowledge_base": [
      "related(x, y)",
      "related(y, z)"
    ],
    "rules": [
      {
        "name": "Transitive Rule",
        "forall": [
          {"name": "a", "sort": "Element"},
          {"name": "b", "sort": "Element"},
          {"name": "c", "sort": "Element"}
        ],
        "implies": {
          "antecedent": "And(related(a, b), related(b, c))",
          "consequent": "related(a, c)"
        }
      }
    ],
    "verifications": [
      {
        "name": "Verify Transitivity",
        "constraint": "related(x, z)"
      }
    ],
    "actions": ["verify_conditions"]
  }

\end{lstlisting}

  \item \textbf{Scheduling Without Conflicts} : Ensure two tasks are scheduled at different times.
    \begin{lstlisting}[basicstyle=\tiny, language=json,firstnumber=1]

{
    "sorts": [
      {"name": "Task", "type": "DeclareSort"},
      {"name": "TimeSlot", "type": "IntSort"}
    ],
    "functions": [
      {"name": "scheduled_at", "domain": ["Task"], "range": "TimeSlot"}
    ],
    "constants": {
      "tasks": {
        "sort": "Task",
        "members": ["task1", "task2"]
      }
    },
    "knowledge_base": [],
    "rules": [],
    "verifications": [
      {
        "name": "Verify Scheduling",
        "exists": [
          {"name": "t1", "sort": "TimeSlot"},
          {"name": "t2", "sort": "TimeSlot"}
        ],
        "constraint": "And(scheduled_at(task1) == t1, scheduled_at(task2) == t2, t1 != t2)"
      }
    ],
    "actions": ["verify_conditions"]
  }
\end{lstlisting}

  \item \textbf{Graph Coloring Problem} :  Assign colors to nodes such that adjacent nodes have different colors.
     \begin{lstlisting}[basicstyle=\tiny, language=json,firstnumber=1]

  {
    "sorts": [
      {"name": "Node", "type": "DeclareSort"},
      {"name": "Color", "type": "DeclareSort"}
    ],
    "functions": [
      {"name": "color_of", "domain": ["Node"], "range": "Color"},
      {"name": "connected", "domain": ["Node", "Node"], "range": "BoolSort"}
    ],
    "constants": {
      "nodes": {
        "sort": "Node",
        "members": ["node1", "node2", "node3"]
      },
      "colors": {
        "sort": "Color",
        "members": ["red", "green", "blue"]
      }
    },
    "knowledge_base": [
      "connected(node1, node2)",
      "connected(node2, node3)",
      "connected(node1, node3)"
    ],
    "rules": [
      {
        "name": "Coloring Rule",
        "forall": [
          {"name": "n1", "sort": "Node"},
          {"name": "n2", "sort": "Node"}
        ],
        "implies": {
          "antecedent": "connected(n1, n2)",
          "consequent": "color_of(n1) != color_of(n2)"
        }
      }
    ],
    "verifications": [
      {
        "name": "Verify Coloring",
        "exists": [
          {"name": "c1", "sort": "Color"},
          {"name": "c2", "sort": "Color"},
          {"name": "c3", "sort": "Color"}
        ],
        "constraint": "And(color_of(node1) == c1, color_of(node2) == c2, color_of(node3) == c3)"
      }
    ],
    "actions": ["verify_conditions"]
  }
\end{lstlisting}

  \item \textbf{Health and Safety Scenario} : Verify that all workers at heights above 6 feet are wearing safety harnesses.
    \begin{lstlisting}[basicstyle=\tiny, language=json,firstnumber=1]

  {
    "sorts": [
      {"name": "Person", "type": "DeclareSort"},
      {"name": "Equipment", "type": "DeclareSort"},
      {"name": "Location", "type": "DeclareSort"}
    ],
    "functions": [
      {"name": "Worker", "domain": ["Person"], "range": "BoolSort"},
      {"name": "At", "domain": ["Person", "Location"], "range": "BoolSort"},
      {"name": "Wearing", "domain": ["Person", "Equipment"], "range": "BoolSort"},
      {"name": "Height", "domain": ["Location"], "range": "IntSort"}
    ],
    "constants": {
      "persons": {
        "sort": "Person",
        "members": ["worker1", "worker2"]
      },
      "equipments": {
        "sort": "Equipment",
        "members": ["safetyHarness"]
      },
      "locations": {
        "sort": "Location",
        "members": ["groundLevel", "highLevel"]
      }
    },
    "knowledge_base": [
      "Worker(worker1)",
      "Worker(worker2)",
      "At(worker1, groundLevel)",
      "At(worker2, highLevel)",
      "Height(groundLevel) == 0",
      "Height(highLevel) == 20",
      "Wearing(worker1, safetyHarness)"
    ],
    "rules": [
      {
        "name": "Fall Protection Rule",
        "forall": [
          {"name": "p", "sort": "Person"},
          {"name": "l", "sort": "Location"}
        ],
        "implies": {
          "antecedent": "And(Worker(p), At(p, l), Height(l) > 6)",
          "consequent": "Wearing(p, safetyHarness)"
        }
      }
    ],
    "verifications": [
      {
        "name": "Check Fall Protection",
        "forall": [
          {"name": "p", "sort": "Person"},
          {"name": "l", "sort": "Location"}
        ],
        "implies": {
          "antecedent": "And(Worker(p), At(p, l), Height(l) > 6)",
          "consequent": "Wearing(p, safetyHarness)"
        }
      }
    ],
    "actions": ["verify_conditions"]
  }
\end{lstlisting}
  
    \item \textbf{Electrical Safety Scenario}: Ensure workers using energized equipment above 250V are wearing insulated gloves.
      \begin{lstlisting}[basicstyle=\tiny, language=json,firstnumber=1]

{
    "sorts": [
      {"name": "Person", "type": "DeclareSort"},
      {"name": "Equipment", "type": "DeclareSort"}
    ],
    "functions": [
      {"name": "Worker", "domain": ["Person"], "range": "BoolSort"},
      {"name": "Using", "domain": ["Person", "Equipment"], "range": "BoolSort"},
      {"name": "IsEnergized", "domain": ["Equipment"], "range": "BoolSort"},
      {"name": "Voltage", "domain": ["Equipment"], "range": "IntSort"},
      {"name": "Wearing", "domain": ["Person", "Equipment"], "range": "BoolSort"}
    ],
    "constants": {
      "persons": {
        "sort": "Person",
        "members": ["worker1"]
      },
      "equipments": {
        "sort": "Equipment",
        "members": ["circuitBreaker", "insulatedGloves"]
      }
    },
    "knowledge_base": [
      "Worker(worker1)",
      "Using(worker1, circuitBreaker)",
      "IsEnergized(circuitBreaker)",
      "Voltage(circuitBreaker) == 480"
    ],
    "rules": [
      {
        "name": "High Voltage Safety Rule",
        "forall": [
          {"name": "p", "sort": "Person"},
          {"name": "e", "sort": "Equipment"}
        ],
        "implies": {
          "antecedent": "And(Worker(p), Using(p, e), IsEnergized(e), Voltage(e) > 250)",
          "consequent": "Wearing(p, insulatedGloves)"
        }
      }
    ],
    "verifications": [
      {
        "name": "Verify Electrical Safety",
        "forall": [
          {"name": "p", "sort": "Person"},
          {"name": "e", "sort": "Equipment"}
        ],
        "implies": {
          "antecedent": "And(Worker(p), Using(p, e), IsEnergized(e), Voltage(e) > 250)",
          "consequent": "Wearing(p, insulatedGloves)"
        }
      }
    ],
    "actions": ["verify_conditions"]
  }
\end{lstlisting}

\item \textbf{Chemical Handling Safety} :  Ensure workers handling corrosive chemicals are wearing gloves and goggles.
    \begin{lstlisting}[basicstyle=\tiny, language=json,firstnumber=1]

{
    "sorts": [
      {"name": "Person", "type": "DeclareSort"},
      {"name": "Chemical", "type": "DeclareSort"},
      {"name": "Equipment", "type": "DeclareSort"}
    ],
    "functions": [
      {"name": "Worker", "domain": ["Person"], "range": "BoolSort"},
      {"name": "Handling", "domain": ["Person", "Chemical"], "range": "BoolSort"},
      {"name": "IsCorrosive", "domain": ["Chemical"], "range": "BoolSort"},
      {"name": "Wearing", "domain": ["Person", "Equipment"], "range": "BoolSort"}
    ],
    "constants": {
      "persons": {
        "sort": "Person",
        "members": ["worker1"]
      },
      "chemicals": {
        "sort": "Chemical",
        "members": ["acid"]
      },
      "equipments": {
        "sort": "Equipment",
        "members": ["gloves", "goggles"]
      }
    },
    "knowledge_base": [
      "Worker(worker1)",
      "Handling(worker1, acid)",
      "IsCorrosive(acid)"
    ],
    "rules": [
      {
        "name": "Corrosive Chemical Handling Rule",
        "forall": [
          {"name": "p", "sort": "Person"},
          {"name": "c", "sort": "Chemical"}
        ],
        "implies": {
          "antecedent": "And(Worker(p), Handling(p, c), IsCorrosive(c))",
          "consequent": "And(Wearing(p, gloves), Wearing(p, goggles))"
        }
      }
    ],
    "verifications": [
      {
        "name": "Verify Chemical Safety",
        "forall": [
          {"name": "p", "sort": "Person"},
          {"name": "c", "sort": "Chemical"}
        ],
        "implies": {
          "antecedent": "And(Worker(p), Handling(p, c), IsCorrosive(c))",
          "consequent": "And(Wearing(p, gloves), Wearing(p, goggles))"
        }
      }
    ],
    "actions": ["verify_conditions"]
  }
\end{lstlisting}

  \item \textbf{Resource Allocation Optimization} : Allocate tasks to workers while minimizing total cost.
     \begin{lstlisting}[basicstyle=\tiny, language=json,firstnumber=1]

  {
    "sorts": [
      {"name": "Worker", "type": "DeclareSort"},
      {"name": "Task", "type": "DeclareSort"},
      {"name": "Cost", "type": "IntSort"}
    ],
    "functions": [
      {"name": "assigned_to", "domain": ["Task"], "range": "Worker"},
      {"name": "cost_of", "domain": ["Worker"], "range": "Cost"}
    ],
    "constants": {
      "workers": {
        "sort": "Worker",
        "members": ["worker1", "worker2"]
      },
      "tasks": {
        "sort": "Task",
        "members": ["taskA", "taskB"]
      }
    },
    "knowledge_base": [
      "cost_of(worker1) == 50",
      "cost_of(worker2) == 30"
    ],
    "rules": [],
    "verifications": [],
    "optimization": {
      "constraints": [
        "assigned_to(taskA) != assigned_to(taskB)"
      ],
      "objectives": [
        {
          "type": "minimize",
          "expression": "cost_of(assigned_to(taskA)) + cost_of(assigned_to(taskB))"
        }
      ]
    },
    "actions": ["optimize"]
  }
  
\end{lstlisting}

\end{enumerate}

\subsection{Exploring the possibilities: Unsatisfiable Neurosymbolic Programs}
Our DSL designed to be very expressive, and future proof for additional scenarios. In this subsection we present some example problems that can be expressed and solved to be found UNSAT.

\begin{enumerate}
    \item \textbf{Pigeonhole Principle:} A classic unsatisfiable problem where more pigeons than holes cannot be assigned uniquely.
    \begin{lstlisting}[basicstyle=\tiny, language=json,firstnumber=1]
{
    "sorts": [
      {
        "name": "Pigeon",
        "type": "EnumSort",
        "values": ["p1", "p2", "p3", "p4"]
      },
      {
        "name": "Hole",
        "type": "EnumSort",
        "values": ["h1", "h2", "h3"]
      }
    ],
    "functions": [
      {"name": "assigned_to", "domain": ["Pigeon"], "range": "Hole"}
    ],
    "constants": {},
    "knowledge_base": [],
    "rules": [],
    "verifications": [
      {
        "name": "Pigeonhole Verification",
        "constraint": "Distinct(assigned_to(p1), assigned_to(p2), assigned_to(p3), assigned_to(p4))"
      }
    ],
    "actions": ["verify_conditions"]
  }
  
\end{lstlisting}

\item \textbf{Non-Three-Colorable Graph :} A complete graph with four nodes cannot be colored with only three colors without adjacent nodes sharing the same color.

\begin{lstlisting}[language=json,firstnumber=1]
{
    "sorts": [
      {
        "name": "Node",
        "type": "EnumSort",
        "values": ["n1", "n2", "n3", "n4"]
      },
      {
        "name": "Color",
        "type": "EnumSort",
        "values": ["red", "green", "blue"]
      }
    ],
    "functions": [
      {"name": "color_of", "domain": ["Node"], "range": "Color"},
      {"name": "connected", "domain": ["Node", "Node"], "range": "BoolSort"}
    ],
    "constants": {},
    "knowledge_base": [
      "connected(n1, n2)",
      "connected(n1, n3)",
      "connected(n1, n4)",
      "connected(n2, n3)",
      "connected(n2, n4)",
      "connected(n3, n4)"
    ],
    "rules": [
      {
        "name": "Coloring Rule",
        "forall": [
          {"name": "n1", "sort": "Node"},
          {"name": "n2", "sort": "Node"}
        ],
        "implies": {
          "antecedent": "And(connected(n1, n2), n1 != n2)",
          "consequent": "color_of(n1) != color_of(n2)"
        }
      }
    ],
    "verifications": [
      {
        "name": "Verify 3-Coloring",
        "constraint": "True"
      }
    ],
    "actions": ["verify_conditions"]
  }
  \end{lstlisting}

\item \textbf{Contradictory Mathematical Constraints}
\begin{lstlisting}[language=json,firstnumber=1]

{
    "sorts": [
      {"name": "Int", "type": "IntSort"}
    ],
    "functions": [],
    "constants": {},
    "knowledge_base": [],
    "rules": [],
    "verifications": [
      {
        "name": "Contradictory Constraints",
        "exists": [
          {"name": "x", "sort": "Int"}
        ],
        "constraint": "And(x > 0, x < 0)"
      }
    ],
    "actions": ["verify_conditions"]
  }
    \end{lstlisting}

\item \textbf{An Unsatisfiable Boolean formula} in conjunctive normal form (CNF).
\begin{lstlisting}[language=json,firstnumber=1]

{
    "sorts": [
      {"name": "Bool", "type": "BoolSort"}
    ],
    "functions": [],
    "constants": {
      "variables": {
        "sort": "Bool",
        "members": ["A", "B"]
      }
    },
    "knowledge_base": [],
    "rules": [],
    "verifications": [
      {
        "name": "Unsatisfiable CNF",
        "constraint": "And(A, Not(A))"
      }
    ],
    "actions": ["verify_conditions"]
  }
  \end{lstlisting}

\item \textbf{Mutual Exclusivity}: Defining two constants that cannot be equal, but also constrained to be equal.
\begin{lstlisting}[language=json,firstnumber=1]

{
    "sorts": [
      {"name": "Element", "type": "DeclareSort"}
    ],
    "functions": [],
    "constants": {
      "elements": {
        "sort": "Element",
        "members": ["e1", "e2"]
      }
    },
    "knowledge_base": [
      "e1 != e2",
      "e1 == e2"
    ],
    "rules": [],
    "verifications": [
      {
        "name": "Mutual Exclusivity Verification",
        "constraint": "True"
      }
    ],
    "actions": ["verify_conditions"]
  }

  \end{lstlisting}

  \item \textbf{Inconsistent Equations}
\begin{lstlisting}[language=json,firstnumber=1]

  {
    "sorts": [
      {"name": "Int", "type": "IntSort"}
    ],
    "functions": [],
    "constants": {},
    "knowledge_base": [],
    "rules": [],
    "verifications": [
      {
        "name": "Inconsistent Equations",
        "exists": [
          {"name": "x", "sort": "Int"},
          {"name": "y", "sort": "Int"}
        ],
        "constraint": "And(x + y == 10, x + y == 5)"
      }
    ],
    "actions": ["verify_conditions"]
  }
  \end{lstlisting}

  \item \textbf{Unsolvable Scheduling Conflict} : Tasks that must occur at the same time and also at different times.
\begin{lstlisting}[language=json,firstnumber=1]

{
    "sorts": [
      {"name": "Task", "type": "EnumSort", "values": ["task1", "task2"]}
    ],
    "functions": [
      {"name": "scheduled_at", "domain": ["Task"], "range": "IntSort"}
    ],
    "constants": {},
    "knowledge_base": [
      "scheduled_at(task1) == scheduled_at(task2)",
      "scheduled_at(task1) != scheduled_at(task2)"
    ],
    "rules": [],
    "verifications": [
      {
        "name": "Scheduling Conflict Verification",
        "constraint": "True"
      }
    ],
    "actions": ["verify_conditions"]
  }
  \end{lstlisting}

  \item \textbf{Invalid Parent-Child Relationship}
\begin{lstlisting}[language=json,firstnumber=1]

{
    "sorts": [
      {"name": "Person", "type": "EnumSort", "values": ["bob"]}
    ],
    "functions": [
      {"name": "parent_of", "domain": ["Person"], "range": "Person"}
    ],
    "constants": {},
    "knowledge_base": [
      "parent_of(bob) == bob"
    ],
    "rules": [
      {
        "name": "No Self Parenting Rule",
        "forall": [
          {"name": "p", "sort": "Person"}
        ],
        "implies": {
          "antecedent": "True",
          "consequent": "parent_of(p) != p"
        }
      }
    ],
    "verifications": [
      {
        "name": "Self Parenting Verification",
        "constraint": "True"
      }
    ],
    "actions": ["verify_conditions"]
  }
  \end{lstlisting}

  \item \textbf{Impossible Optimization}
\begin{lstlisting}[language=json,firstnumber=1]

  {
  "sorts": [
    {"name": "Int", "type": "IntSort"}
  ],
  "functions": [],
  "constants": {},
  "knowledge_base": [],
  "rules": [],
  "verifications": [],
  "optimization": {
    "constraints": [
      "x > 0",
      "x < 0"
    ],
    "objectives": [
      {
        "type": "minimize",
        "expression": "x"
      }
    ]
  },
  "actions": ["optimize"]
}
  \end{lstlisting}
\end{enumerate}